\documentclass[]{article}

\usepackage{arxiv}

\usepackage{bm}
\usepackage{amsmath}
\usepackage{amssymb}
\usepackage{amsthm}
\usepackage{bbm}
\usepackage{graphicx}
\usepackage[
    backend=biber,
    hyperref = true,
    bibstyle=authoryear-icomp,
    citestyle=authoryear-icomp,
    url=false,
]{biblatex}
\usepackage{url}
\usepackage[
  luatex
]{hyperref}
\usepackage{cleveref}
\usepackage{tcolorbox}
\usepackage{booktabs}
\usepackage{subcaption}
\usepackage{csquotes}
\usepackage[
  obeyDraft
]{todonotes}
\usepackage{framed}
\usepackage{authblk}
\usepackage{orcidlink}
\usepackage[utf8]{inputenc} 
\usepackage[T1]{fontenc}    
\usepackage{amsfonts}       
\usepackage{nicefrac}       
\usepackage{microtype}      
\usepackage{doi}

\theoremstyle{plain}

\crefname{equation}{Equation}{Equations}
\crefname{table}{Table}{Tables}
\crefname{figure}{Figure}{Figures}
\crefname{section}{Section}{Sections}
\crefname{theorem}{Theorem}{Theorems}
\crefname{corollary}{Corollary}{Corollaries}

\DeclareMathOperator*{\concat}{%
    \mathchoice%
        {\Big\Vert}%
        {\big\Vert}%
        {\Vert}%
        {\Vert}%
}

\title{Do Large Language Models Advocate for\\Inferentialism?}

\renewcommand{\shorttitle}{Do Large Language Models Advocate for Inferentialism?}

\hypersetup{
pdftitle={Do Large Language Models Advocate for Inferentialism?},
pdfsubject={cs.CL},
pdfauthor={Yuzuki Arai, Sho Tsugawa},
pdfkeywords={large language models, philosophy of language, inferentialism}
}

\date{}

\author[1]{Yuzuki Arai\,\orcidlink{0000-0002-0333-6103}}
\author[2]{Sho Tsugawa\,\orcidlink{0000-0001-7837-2857}}

\affil[1]{College of Media Arts, Science and Technology, School of Informatics, University of Tsukuba\thanks{\texttt{y-arai@snlab.cs.tsukuba.ac.jp}}}
\affil[2]{Institute of Systems and Information Engineering, University of Tsukuba\thanks{\texttt{s-tugawa@cs.tsukuba.ac.jp}}}

\begin{document}

\maketitle

\begin{abstract}
The emergence of large language models (LLMs) such as ChatGPT and Claude presents new challenges for philosophy of language, particularly regarding the nature of linguistic meaning and representation. While LLMs have traditionally been understood through distributional semantics, this paper explores Robert Brandom's inferential semantics as an alternative foundational framework for understanding these systems. We examine how key features of inferential semantics---including its anti-representationalist stance, logical expressivism, and quasi-compositional approach---align with the architectural and functional characteristics of Transformer-based LLMs. Through analysis of the ISA (Inference, Substitution, Anaphora) approach, we demonstrate that LLMs exhibit fundamentally anti-representationalist properties in their processing of language. We further develop a consensus theory of truth appropriate for LLMs, grounded in their interactive and normative dimensions through mechanisms like RLHF. While acknowledging significant tensions between inferentialism's philosophical commitments and LLMs' sub-symbolic processing, this paper argues that inferential semantics provides valuable insights into how LLMs generate meaning without reference to external world representations. Our analysis suggests that LLMs may challenge traditional assumptions in philosophy of language, including strict compositionality and semantic externalism, though further empirical investigation is needed to fully substantiate these theoretical claims.
\end{abstract}


\section{Introduction}

The emergence of large language models (LLMs) such as ChatGPT (OpenAI), Claude (Anthropic), Gemini (Google), and Microsoft Copilot (Microsoft) presents new challenges for philosophy of language, particularly regarding the nature of linguistic meaning and representation \autocite{Cappelen2021-CAPMAI,milliere2024philosophicalintroductionlanguagemodels,Nefdt_2024}. Traditionally, LLMs have been explained through distributional semantics as their foundational semantics \autocite{grindrod2024large, enyan2024llmsmodelsdistributionalsemantics, Lenci_Sahlgren_2023}. Recent research, however, explores alternative foundational semantics beyond distributional semantics \autocite{grindrod2024large, Mallory2023Nov}. This paper proposes Robert Brandom's inferential semantics \autocite{Brandom1994-BRAMIE} as an suitable foundational semantics for LLMs, focusing specifically on the issue of linguistic representationalism within the post-anthropocentric paradigm of philosophy of language.

In philosophical inquiry into the nature of language and the process of meaning formation, truth-conditional semantics and inferential semantics have stood as two major opposing approaches \autocite[chap.5.2]{weiss_wanderer_2010}. The former aligns itself with representationalism, while the latter champions anti\hyphen{}representationalism. This debate represents a fundamental conflict within the philosophy of language and epistemology, offering fundamentally different views on the relationship between language and the world.

Representationalism views language as a mirror reflecting truths about the world. Truth-conditional semantics, a leading representationalist approach, defines the meaning of a sentence by its truth conditions—that is, the necessary and sufficient conditions under which it is true \autocite{Heim1998-HEISIG}. A key example of this, notably applied in engineering fields such as natural language processing, is Montague semantics, a model-theoretic formal semantics grounded in logic that assumes compositionality in meaning and defines meaning through models of possible worlds, entities, and universals \autocite{Partee_2016}.

Conversely, anti\hyphen{}representationalism eschews external representations from the outset to circumvent the skepticism surrounding meaning, which posits that there are no facts that ground the meaning of a word or sentence \autocite{soames_skepticism_1997,Kripke1982-KRIWOR}. This approach, often termed anti\hyphen{}representationalism, can be traced back to Richard Rorty's seminal work, \citetitle{Rorty1979-RORPAT-3}. One particular form of anti\hyphen{}representationalist semantics is the inferential semantics proposed by \textcite{Brandom1994-BRAMIE}, which grounds meaning not in external representations but in the norms and inferential roles among language users.

This paper suggests that the anti-representational nature of LLMs may align with Robert Brandom's inferential semantics \autocite{Brandom1994-BRAMIE}, potentially offering an alternative perspective for understanding their functions and characteristics in comparison to distributional semantics or the widely accepted truth-conditional semantics in contemporary philosophy of language. While the merits of inferential semantics and inferential role semantics as foundational semantics for LLMs have already been suggested by \textcite{havlik_meaning_2024} and \textcite{borg_llms_nodate}, this paper aims to explore this possibility more systematically from the vantage point of LLM architecture.

The structure of the paper is as follows. First, we provide a brief explanation of LLMs and the paradigm they belong to (\cref{sec:llm}). Next, we argue for the suitability of inferential semantics over distributional semantics as the foundational semantics for LLMs (\cref{sec:semantics-of-llms}). We then outline the philosophical debate between representationalism and anti\hyphen{}representationalism and discuss how inferential semantics falls within the latter category (\cref{sec:representationalism}). Following this, we explain the core concepts of the ISA approach and linguistic idealism within inferential semantics (\cref{sec:inferentialism}). We further analyze the nature of LLMs through the three components of the ISA approach: inference, substitution, and anaphora, demonstrating that inferentialism's anti\hyphen{}representationalism aligns with the characteristics of LLMs (\cref{sec:isa-and-llm,sec:llm_anti_representationalism}). Next, we propose a consensus theory of truth for LLMs and discuss normativity (\cref{sec:llm_truth}). Finally, we examine the limitations of interpreting LLMs through inferentialism (\cref{sec:tensions}).

This paper argues that the properties of LLMs challenge mainstream assumptions in the philosophy of language, such as semantic externalism and compositionality. We believe that the argument in this paper leads to a reevaluation of anti\hyphen{}representationalist views of language, potentially leading to new developments in inferentialism and the philosophy of language.

\section{Large Language Models}\label{sec:llm}

LLMs such as OpenAI's GPT-4, Anthropic's Claude Opus, Meta's Llama 3, and InstructGPT \autocite{ouyang2022traininglanguagemodelsfollow} are based on an architecture called Transformer, proposed by \textcite{NIPS2017_3f5ee243} \footnote{The arguments about LLMs in this paper also apply to small language models (SLMs) \autocite{lu2024smalllanguagemodelssurvey, vannguyen2024surveysmalllanguagemodels}. Small language models refer to language models that are designed to achieve performance comparable performance to LLMs while using fewer parameters and less computational resources.}. In particular, Transformers with multiple heads and layers are referred to as multi-head, multi-layer Transformers\footnote{For a detailed mathematical explanation of the Transformer architecture, see \textcite{elhage2021mathematical}.}. The multi-head, multi-layer Transformer, a central component of LLMs, retains information separated across subspaces corresponding to each head, allowing the model to represent syntactic structures, categorical hierarchies, and concepts \autocite{voita-etal-2019-analyzing}. The Transformer takes as input the sum of embedding vectors (words or characters) and positional vectors ($\boldsymbol{v}_i + \boldsymbol{p}_i$), and computes self-attention through the QKV mechanism within the Transformer. In this process, the relationships between elements of the input sequence are captured. The vectors flowing through the residual stream undergo linear transformations in the attention mechanism, producing three vectors: queries ($Q$), keys ($K$), and values ($V$)\footnote{
In \cref{eq:transformer_1}, the concatenation operator ($\concat$) joins given vectors or matrices sequentially.
This operation corresponds to the direct sum of vector spaces.
Conversely, splitting vectors (for example, splitting a 4-dimensional vector $(x, y, z, w)$ into two 2-dimensional vectors $(x, y)$ and $(z, w)$) corresponds to the direct sum decomposition of vector spaces.
Both are operations between subspaces that are mutually orthogonal complements, enabling the multi-head Transformer architecture to retain information independently.
}. \cref{eq:transformer_1,eq:transformer_2,eq:transformer_3} are the mathematical representation of a typical multi-head Transformer architecture. \cref{eq:transformer_1} shows that the geometric function of multi-head attention lies in the direct sum of multiple subspaces. \cref{eq:transformer_2} indicates that the core of the self-attention mechanism (QKV mechanism) is the dot product between the query vector ($Q$) and the key vector ($K$). \cref{eq:transformer_3} shows that the self-attention operation is performed for each head.


\begin{align}
    \operatorname{MultiHead}(Q, K, V) = \concat_{i=1}^h H_i W^O \label{eq:transformer_1} \\
    \operatorname{Attn}(Q, K, V) = \operatorname{softmax}\left(\frac{QK^T}{\sqrt{d_k}}\right)V \label{eq:transformer_2} \\
    H_i = \operatorname{Attn}(QW_i^Q, KW_i^K, VW_i^V)
    \label{eq:transformer_3}
\end{align}

The orthogonality of subspaces defined by the weight matrices ($W$) performing linear transformations enables information to be retained separately within each layer and each head of the model. This multi-layer, multi-head structure allows the model to acquire hierarchical and distributed representations, enabling it to capture complex linguistic phenomena. The Transformer architecture can address the intricate syntactic properties of language by capturing word-level features in shallow layers and abstract concepts or context in deeper layers.

The multi-head, multi-layer Transformers combine properties of classical symbolicism and classical connectionism. Symbolicism (computationalism) views human thought as a chain of discrete symbolic representations. In contrast, connectionism assumes internal representations are distributed within neural networks\footnote{The difference between symbolicism and connectionism can also be seen in the tools they use: the former relies on logic, while the latter employs linear algebra, calculus, and topology.}. Traditional AI before LLMs, known as GOFAI (Good Old-Fashioned Artificial Intelligence), belonged to the symbolicism paradigm, aiming to process natural language through discrete symbol manipulation. In contrast, LLMs are based on artificial neural network architectures, belonging to the connectionist paradigm while also exhibiting symbolic behavior in higher layers, such as syntax and coreference resolution\footnote{For a comparison between GOFAI and LLMs, see \textcite{Gubelmann2023-GUBALW}.}. \textcite{Chalmers2023-CHATCA-21} argues that the properties of LLMs fall under what \autocite{Smolensky1987Jun} refers to as \textit{subsymbolism}.

In addition to the inherent features of the mechanisms described above, contemporary conversational LLMs also employ a mechanism called Human in the loop reinforcement learning (RLHF)\footnote{For technical details on RLHF, see \textcite{ouyang2022traininglanguagemodelsfollow, 10.5555/3294996.3295184}.}. RLHF utilizes human evaluations of model outputs as reward signals to adjust the model's behavior according to human expectations.
RLHF consists of three main stages. First, human annotators evaluate multiple responses to given prompts, ranking them by preference. Second, a reward model is trained using these ranking data. Finally, reinforcement learning is employed with this reward model to adjust the LLM to generate outputs that receive high rewards.
Mathematically, the training of the reward model can be formulated as minimizing the following loss function:

\begin{align}
L(\theta) = -\mathbb{E}[\ln(\sigma(r_\theta(x,y_w) - r_\theta(x,y_l)))]
\end{align}

Here, $r_\theta$ is the reward model with parameters $\theta$, $x$ is the input prompt, $y_w$ is the response preferred by humans, $y_l$ is the less preferred response, and $\sigma$ is the sigmoid function. This equation represents the training of the reward model to appropriately distinguish between preferred and non-preferred responses.
In the reinforcement learning stage, the following objective function is maximized:

\begin{align}
\text{objective}(\phi) = \mathbb{E}[r_\theta(x,y) - \beta \cdot \text{KL}(\pi_\phi^{\text{RL}}(y|x) | \pi^{\text{IT}}(y|x))]
\end{align}

In this equation, $\pi_\phi^{\text{RL}}$ is the policy optimized by reinforcement learning (the LLM after RL), $\pi^{\text{IT}}$ is the initial policy after instruction tuning, and $\beta$ is the weight coefficient for the KL divergence term. This objective function maximizes rewards while constraining the model from deviating excessively from the initial policy.
Instruction tuning for LLMs is a learning process that enhances the ability to follow specific instructions. This combines supervised fine-tuning (SFT), which provides correct answers based on specific task instructions, with the aforementioned RLHF. This two-stage process enables LLMs to generate responses that follow explicit instructions while aligning with human values and preferences. For example, behaviors such as avoiding harmful content generation and providing useful information are reinforced.

Another feedback method, known as Direct Preference Optimization (DPO), has also been proposed. DPO presents an alternative to traditional reinforcement learning-based methods, such as RLHF. DPO does not employ the training of a discrete reward model for policy optimization; rather, it utilizes a direct fine-tuning of the model based on human preference data. The learning process is conceptualized as a binary classification problem between preferred and dispreferred responses, and the model is optimized by minimizing a contrastive loss function. This approach simplifies the pipeline by removing the need for reward modeling and reinforcement learning, while still effectively aligning model outputs with human preferences \autocite{10.5555/3666122.3668460}.

\section{Foundational Semantics of Large Language Models}\label{sec:semantics-of-llms}

In the philosophy of language, \textit{semantics} is divided into two branches: foundational semantics and descriptive semantics \autocite[535]{Stalnaker1997-STARAN-2}\footnote{Foundational semantics is also called meta-semantics, while descriptive semantics is sometimes referred to simply as semantics \autocite[pp.~573--4]{Kaplan1989-KAPA}.}. Foundational semantics is a theory that explains the very meaning of meaning itself and includes truth-conditional semantics, verificationist semantics \autocite{Dummett1975-DUMWIA-3}, inferential semantics \autocite{Brandom1994-BRAMIE}, and teleosemantics \autocite{Millikan1984-MILLTA-3}. In contrast, descriptive semantics refers to theories that assign semantic value to sentences within a given semantic framework. Descriptive semantics bears a resemblance to the semantics described in the division of linguistic inquiry into pragmatics, semantics, and syntax by Charles Morris \autocite{posner_charles_1987}.

The prevailing foundational semantics most often attributed to LLMs is distributional semantics, which is constructed within the American structuralist linguistic tradition by \textcite{firth1957synopsis} and \textcite{doi:10.1080/00437956.1954.11659520} \autocite{grindrod2024large}. Distributional semantics posits the notion that the meaning of a lexical item is determined by the distribution of its surrounding contexts \autocite{Grindrod2023-GRIDTO-2}. On the question of whether distributional semantics can encompass LLMs, \citeauthor{grindrod2024large} states:

\begin{displayquote}[{\cite{grindrod2024large}}]
The crucial linguistic point to note about transformers is that the architecture does not depart from the basic distributional approach insofar as they take the word embeddings that are at the heart of the distributional approach and then modify them on the basis of their contexts, where the context itself is only represented in distributional terms.
\end{displayquote}


It is important to note the distinction between \textit{static embedding models} and \textit{dynamic embedding models} here. Static embedding models such as Word2vec \autocite{mikolov2013efficientestimationwordrepresentations} and GloVe \autocite{pennington-etal-2014-glove} operationalize distributional semantics via explicit co-occurrence statistics of words in text corpora. In contrast to these static models, dynamic embedding models, particularly Transformer-based models such as GPT and BERT, learn contextual word representations not through explicit word co-occurrence probabilities but by optimizing next-token prediction over massive corpora using deep neural architectures. While traditional Distributional Semantics Models (DSMs) rely on explicit co-occurrence statistics, modern LLMs implement distributional semantics through a predictive approach: instead of directly computing co-occurrence probabilities, they learn representations by predicting subsequent tokens in sequences\footnote{Some state-of-the-art models employ Byte-Level BPE (Byte Pair Encoding) \autocite{Wang_Cho_Gu_2020} for embeddings. This method operates at the byte level as opposed to the character, word, or sentence levels by repeatedly merging frequent pairs of characters or strings and treating them as new tokens. The implications of this embedding method for the metaphysics of meaning remain an open question.}. Additionally, regarding the additive compositionality of word vectors, such as the famous equation $\vec{king} + \vec{woman} = \vec{queen} + \vec{man}$ demonstrated by \textcite{mikolov-etal-2013-linguistic}, experimental studies have reported that dynamic embedding models show lower accuracy compared to traditional static embedding models \autocite{naito2022revisitingadditivecompositionalityand}.

From this perspective, distributional semantics is regarded as insufficient for elucidating the behavior of LLMs. Furthermore, distributional semantics is a theory of meaning that generally eschews metaphysical aspects, instead only positing that ``words's meaning is its distribution.'' To the best of our knowledge, there is an absence of descriptive semantics within distributional semantics, and the theory provides scant to no explanation of its philosophical implications. Existing critiques of distributional semantics as an inadequate theoretical framework for interpreting LLMs have also emerged in the field of computational linguistics, as evidenced by the work of \textcite{enyan2024llmsmodelsdistributionalsemantics}.

The objective of this study is not to refine distributional semantics; rather, we propose to adopt inferential semantics as their foundational semantics. We understand that analyzing LLMs through inferential semantics is more useful than distributional semantics in elucidating their characteristics, as demonstrated in this paper. Contemporary philosophy of language acknowledges multiple foundational semantics for both natural and artificial languages (e.g., truth-conditional semantics and teleosemantics), with each foundational semantics playing a role in explaining different aspects and scopes of language. The restriction of foundational semantics to distributional semantics goes hand-in-hand with a reductionist view that Transformers merely perform (non-)linear transformations of vectors. This risks undervaluing the significance of LLMs (leading to underclaiming'' \autocite{bowman-2022-dangers} or the Redescription Fallacy'' \autocite[chap.3]{milliere2024philosophicalintroductionlanguagemodels}) and undermining appropriate analyses. It should be noted that the considerations presented herein could alternatively be viewed as an extension of distributional semantics. However, whether our proposal is regarded as an extension of distributional semantics or as a pure application of inferential semantics to LLMs is not our primary concern.

This paper examines how the adoption of inferential semantics as the foundational semantics for LLMs, as opposed to distributional semantics, affects their capacity to explain behavior and properties. It is acknowledged that alternative foundational semantics could be utilised for the interpretation of LLMs. One such alternative is truth-conditional semantics. However, it is important to note that the truth in LLMs is relative to the training data, suggesting that these models generate and interpret sentences independently of the truth conditions defined as \textit{meaning} in truth-conditional semantics\footnote{\textcite[chap.5]{10.1145/3624724} states: ``The model itself has no notion of truth or falsehood because it lacks the means to exercise these concepts in anything like the way we do.''}. Adopting truth-conditional semantics would imply that LLMs generate language devoid of \textit{meaning}, a notion that appears counterintuitive. Consequently, truth-conditional semantics should be rejected when analysing LLMs\footnote{The validity of this argument is contingent upon the theory of truth that underlies the truth conditions. For instance, if the truth in truth conditions is defined by anti-realist truth theories such as the consensus theory (see \cref{sec:llm_truth}) or the coherence theory, it could be argued that large language model outputs are based on truth conditions. In this case, we are examining truth conditions concerning truths about the world external to LLMs. It is important to note that many truth theories rely on propositional beliefs, which raises the question of whether LLMs maintain beliefs in propositional form. This issue requires further philosophical insight to understand the relationship between LLMs and truth.}.

When applying anthropocentric meta-semantics \autocite{Cappelen2021-CAPMAI} to LLMs, the primary concern is whether concepts like intention can be ascribed to them. For instance, the implementation of intention-based agent semantics such as \textcite{Grice1957-GRIM} necessitates the investigation of addressing whether LLMs possess the necessary intentions or minds that support such intentions. Another example is Searle's ``Connection Principle,'' \autocite{Searle1992-SEATRO} which holds that intentions are ontologically dependent on consciousness---we would have to grapple with the complex question of whether consciousness resides in LLMs. However, it should be noted that Brandom's inferentialism remains faithful to Sellars' psychological nominalism, and prioritizes notions of normative-linguistic content over the notions of mental content\footnote{Intuitively, LLMs do not possess psychological intentions. For attempts to attribute intentions, beliefs, or attitudes to LLMs from the interpretivist perspective in the philosophy of mind, see \textcite{10.1162/tacl_a_00690}. For arguments claiming that LLMs exhibit Kantian cognitive autonomy, see \textcite{Gubelmann2024-GUBLLM}.}. Therefore, we can remain neutral regarding ontological issues concerning mental content in LLMs, and the application of inferential semantics to LLMs, which intuitively lack psychological intentions, is valid and should not pose any significant challenges.

\section{Representationalism and Anti\hyphen{}Representationalism}\label{sec:representationalism}

This section provides an overview of the two opposing positions in the philosophy of language: representationalism and anti\hyphen{}representationalism. The inferentialism adopted in this paper as a framework for interpreting LLMs is consistent with anti\hyphen{}representationalism.

\subsection{Representationalism}

Representationalism is the view that language serves to represent or mirror the facts of the world. Since ancient Greek philosophy, the dominant perspective has been that our language and beliefs should be understood in relation to the truth of the world.

As Aristotle states in \textit{Metaphysics} (1011b25), ``to say of what is that it is not, or of what is not that it is, is false, while to say of what is that it is, and of what is not that it is not, is true'' ---a proposition that represents the correspondence theory of truth, which holds that a sentence or proposition is true if it corresponds to a fact in the world. Tarski's definition of truth formalizes Aristotle's correspondence theory and is known as the Convention T: ``The sentence `P' is true if and only if P'' \autocite{0ea4add0-94e6-3877-bd4e-0c67171ac249}. This biconditional links language and metaphysical ontology within truth-conditional semantics.

The transcendental theory of language, which is presented in the early Wittgenstein's \citetitle{Wittgenstein2014-WITTLX}, which brought about the linguistic turn, asserts the correspondence between language and the logical form of the world (``die logische Form''). This theory equates the limits of language with the limits of the world (though his conception of the world is idealistic). The early Wittgenstein's theory is known as the `picture theory', which is best expressed by the following proposition:

\begin{displayquote}[{\cite[Proposition 4.01]{Wittgenstein2014-WITTLX}}]
A proposition is a picture of reality. A proposition is a model of reality as we imagine it.
\end{displayquote}

The linguistic turn in philosophy denotes a paradigm shift in the field, positing that philosophical quandaries emanate from the medium of language itself. This assertion necessitates a rigorous examination of language to unravel these intricacies \autocite{Rorty1967-RORTLT}. Logical positivism, analytic philosophy, and the truth-conditional semantics that dominate in contemporary philosophy of language inherit this representationalist relationship between language and the world. According to \textcite[575]{ade8c13a-e71a-3d39-90b2-98fdfae41321}, ``the central semantic fact about language is that it carries information about the world.''



\subsection{Anti-Representationalism}\label{sec:anti-representationalism}

Anti\hyphen{}representationalism opposes representationalism's view that language is career of the world facts. This position was advocated by Richard Rorty in his book \citetitle{Rorty1979-RORPAT-3} published in \citeyear{Rorty1979-RORPAT-3}. Rorty's anti\hyphen{}representationalism builds on Sellars' critique of the ``myth of the given,'' which argues that epistemic justification occurs within the ``space of reasons'' \autocite[§36]{Sellars1997-SELEAT-8} and that epistemic linguistic practices are normative. In this view, epistemology and semantics are inseparable, and pure, unmediated knowledge of the world is impossible. Consequently, facts and values are inextricably linked. This perspective on truth refutes the dualism that separates epistemology from ontology. Given that truth-conditional semantics relies on this dualism, the contrast between representationalist truth-conditional semantics and anti\hyphen{}representationalist inferentialism becomes clear.

Inferentialism, as delineated by Brandom, is influenced by the subsequent Wittgenstein's use theory of meaning. The use theory of meaning is encapsulated by the following proposition:

\begin{displayquote}[{\cite[§43]{Wittgenstein1953-WITPI-4}}] The meaning of a word is its use in the language.
\end{displayquote}

\noindent
According to Wittgenstein's philosophical perspective, language is regarded as one of numerous forms of social interaction, with the engagement in linguistic activities being analogous to the participation in a specific linguistic game\footnote{Wittgenstein's conclusion that language use is ``bare'' and that scientific investigations into meaning are impossible \autocite[§§109,126]{Wittgenstein1953-WITPI-4} is a provocative assertion. However, when considering computational linguistic paradigms such as BERTology \autocite{Rogers_2020}, mechanistic interpretability \autocite{Kastner2024}, and explainable AI (XAI) \autocite{arrieta_explainable_2020}, it becomes disputable whether strict nihilism about meaning is appropriate. In this regard, \textcite[sect.3]{milliere2024languagemodelsmodelslanguage} offers a taxonomy of computational linguistics methods, categorizing them into behavioral studies, probing studies, and interventional studies. The latter two categories causal relationships between input and output by decoding or intervening in neurons, a practice that appears to contradict Wittgenstein's perspective.}. In addition to him, philosophers of language such as Donald Davidson also belong to this anti-representationalist tradition.

According to the functional linguist \textcite{harder1995}, ``semantics is frozen pragmatics.'' The meaning of linguistic elements is solidified through their recursive use. Zellig Harris, a leading figure in distributionalism, states that ``[l]inguistic elements are identified by the distributions of the contexts in which they appear'' \autocite[chap.1.1.1]{Lenci_Sahlgren_2023}.

Leonard Bloomfield, another structural linguist, similarly notes:

\begin{displayquote}[{\cite[162]{bloomfield1933language}}] There is nothing in the structure of morphemes like \textit{wolf}, \textit{fox}, and \textit{dog} to tell us the relation between their meanings; this [the meaning of words] is a problem for the zoölogist. The zoölogist's definition of these meanings is welcome to us as a practical help, but it cannot be confirmed or rejected on the basis of our science.
\end{displayquote}

\noindent
For analytical philosophers who subscribe to the notion that truth conditions constitute meaning, this tradition of American structural linguistics poses a significant challenge (``[s]emantics with no treatment of truth conditions is not semantic'', \cite[18]{lewis_general_1970}), and it is at times even designated as ``anti-semantic'' \autocite{Murphy_2003}.

This perspective on meaning is consistent with the subsequent philosophical standpoint espoused by Wittgenstein, who placed greater emphasis on description in lieu of analysis. Proponents of the Augustinian perspective, which posits that words serve to name and refer to objects, are designated as `nomenclaturists.' In contrast, those who subscribe to anti\hyphen{}representationalist views, exemplified by the later Wittgenstein and structuralists are termed `anti\hyphen{}nomenclaturists'. The conflict between early and late Wittgenstein can be seen as a reflection of the broader conflict between nomenclaturism and anti-nomenclaturism, as well as between representationalism and anti\hyphen{}representationalism. This theoretical divide manifests as the opposition between formal semantics and structural linguistics. \citeauthor{brandom2000} captures this divide succinctly:

\begin{displayquote}[{\cite[chap.1.2]{brandom2000}}]
    I actually think that the division of pre-Kantian philosophers into representationalists and inferentialists cuts according to deeper principles of their thought than does the nearly coextensional division of them into empiricists and rationalists.
\end{displayquote}

\section{Inferentialism}\label{sec:inferentialism}

Inferentialism is an anti\hyphen{}representationalist theory in the philosophy of language, proposed by Robert Brandom in his book \citetitle{Brandom1994-BRAMIE}. It primarily consists of inferential semantics and normative pragmatics. In this section, we will discuss the components of inferentialism, focusing particularly on the ISA approach within inferential semantics, its anti\hyphen{}representationalist nature, and its linguistic idealism. These elements of inferentialism will serve as a framework for interpreting the mechanisms and behaviors of LLMs in \cref{sec:isa-and-llm}.

\subsection{Inferential Semantics}

Brandom distinguishes between three forms of inferentialism: `weak inferentialism', `strong inferentialism,' and `hyper-inferentialism' \autocite[pp.131--2]{Brandom1994-BRAMIE}. He adopts strong inferentialism for his own theory. Strong inferentialism recognizes not only inferences between conceptual contents but also the situations and consequences related to them, considering inferential articulation sufficient for determining meaning. Inferential articulation refers to making explicit the role that words or sentences play within inference\footnote{The difference between hyper-inferentialism and strong inferentialism lies in the former's exclusion of perceptual experiences from inference, whereas the latter can include them. Purely linguistic LLMs, which rely solely on linguistic data, may align with hyper-inferentialism but not with strong inferentialism. In contrast, multimodal language models (MLLMs) can potentially align with strong inferentialism. Although this study focuses on pure LLMs, the discussion can extend to multimodal models. Therefore, the difference between strong and hyper-inferentialism does not pose a problem for this analysis.}.

A key concept in this framework is `material inference.' In contrast to formal inference, as seen in model-theoretic truth-conditional semantics, material inference reflects our actual inferential practices and social norms. For example, consider the proposition ``$p$'' (``I have twenty dollars'') and the inference to ``$s$'' (``I can buy a paperback priced at twenty dollars''). In classical logic, this inference is invalid because, for ``$s$'' to be true, other propositions like ``$q$'' (``There is a bookstore that sells paperbacks'') and propositions regarding proximity to the bookstore or transportation means must also be true. Formally, this is expressed by the invalid entailment ``$p \land q \land \dots \implies s \vdash p \implies s$.''  
Material inference, however, allows us to treat such inferences as valid insofar as they are endorsed within the discursive space, reflecting our inferential practices. This view, which considers formal inference as a derivative of language use, is called `logical expressivism.'

Furthermore, Brandom's inferentialism introduces the concepts of substitutable expressions and anaphoric expressions. In truth-conditional semantics, concepts like \textit{truth} and \textit{reference} are central. Inferentialism treats these as substitutable expressions and anaphors, respectively. For instance, the concept of truth is treated as an anaphor in statements like ```P' is true,'' where ``true'' anaphorically refers to the expression ```P'.'' Similarly, reference is treated as an anaphor referring to designators within a sentence.

Singular terms and predicates, which plays key roles in representing individuals and universals in Fregean semantics, are given as follows. In inferentialism, these are defined as follows: If the inference from ``$Qa$'' to ``$Qb$'' is valid, and the inference from ``$Qb$'' to ``$Qa$'' is also valid, then ``$a$'' and ``$b$'' are singular terms. Conversely, if the inference from ``$Qa$'' to ``$Q'a$'' is valid but the reverse inference is not, then ``$Q$'' and ``$Q'$'' are predicates.  
For example, the inference from ``The \textit{Morgenstern} is a planet composed primarily of carbon dioxide'' to ``The \textit{Abendstern} is a planet composed primarily of carbon dioxide'' is valid, and the reverse inference is also valid\footnote{Both \textit{Morgenstern} and \textit{Abendstern} refer to Venus. This example is commonly used to illustrate Frege's distinction between ``reference (Bedeutung)'' and ``sense (Sinn).'' Note that substitution inference does not require actual reference to entities.}. Therefore, \textit{Morgenstern} and \textit{Abendstern} are singular terms referring to the same object.  
Such substitutional inferences determine whether elements are singular terms (if symmetric) or predicates (if asymmetric).

In inferential semantics, demonstratives like ``this book'' and ``that'', which were considered as true proper names in the final stage of Russell's theory \autocite{Russell1940-RUSTPO-55}, are treated as anaphors. For instance, in the sentence ``I am reading \textit{Introduction to Inferentialism}, and that book is already out of stock,'' ``that book'' anaphorically refers to ``\textit{Introduction to Inferentialism}.''

Brandom refers to this inferential semantic approach as the ISA approach, derived from the initials of \emph{Inference}, \emph{Substitution}, and \emph{Anaphora}. The ISA approach can be seen as translating representationalist semantic concepts into anti\hyphen{}representationalist semantics. Through the ISA approach, sentences and words connect to the external world, thereby acquiring representationality.

\subsection{Linguistic Idealism}

Inferentialism is sometimes characterized as linguistic idealism. Linguistic idealism is the view that the world constructed by language is all that exists. In inferential semantics, the meanings of words are determined solely by their roles in inference. Therefore, meaning is confined within language, and sentences do not structurally correspond to the world\footnote{Compare with \autocite[Proposition 5.5562]{Wittgenstein2014-WITTLX}, regarding the relationship between ordinary language and the world.}.

Inferentialism has been criticized as a relativist theory that ``loses the world,'' making it unable to describe scientific progress as outlined by Thomas Kuhn \autocite[pp.~330--1]{Brandom1994-BRAMIE} \autocite{e24ccbf1-9721-3ad0-a80c-633a91acaad8,c449437e-74f7-3443-83d8-a8e469c30100,Bernstein2010-BERTPT-9}.  
In response, Brandom argues that introducing the distinction between normative attitudes and normative statuses can avoid relativism \autocite[pp.~54-55, 597]{Brandom1994-BRAMIE}.  
Additionally, in \textcite{Brandom2019-BRAASO-13}, he incorporates conceptual realism to account for world-responsiveness within inferentialism.  
Conceptual realism holds that the world has a conceptual structure that exists independently of us.

\section{Large Language Models and the ISA Approach}\label{sec:isa-and-llm}

In this section, we will discuss LLMs and their connection to the inferentialist ISA approach, showing how the characteristics of LLMs support the logical expressivism and quasi-compositional, anti-representational principles of inferentialism.

\subsection{Inference}

In classical symbolist approaches, formal inference constituted inference itself, with its syntax and semantics described by logic. In contrast, the Transformer architecture performs inference not through formal logic, but through statistical rules captured by the heads in higher layers. LLMs acquire inference capabilities from patterns of language use contained in the training data, without being explicitly given logical inference rules. This feature fits well with the concept of material inference and is less compatible with truth-conditional semantics, which depend on formal inference.

\textcite{ae1d0a0a-8dea-323c-89d0-72c600931595} defends the indispensability of material inference to our linguistic practices, exemplified by the inference ``A is an apple $\implies$ A is a fruit'' (MI). Proponents who emphasize formal inference consider this inference as an enthymematic form of a syllogism and do not accept it in its raw form. They regard inference (MI) as an abbreviated form of the following syllogism:

\begin{quote}
    \textbf{Premise 1}: A is an apple \\
    \textbf{Premise 2}: For all X, if X is an apple, then X is a fruit \\
    \textbf{Conclusion}: A is a fruit
\end{quote}

The inference capability of LLMs is trained through the material use of inference rules embedded in their training data. It is then stored in the weight matrices of the higher layers of the model and expressed when these weight matrices operate during token output. LLMs do not perform inference by some indirect means such as analyzing truncated syllogisms (enthymemes) in formal inference. The material inferences advocated by \textcite{ae1d0a0a-8dea-323c-89d0-72c600931595} are derived directly from statistical patterns whenever such emergent sufficient inference rules are included in the training data. Thus, it is appropriate to think of inference in LLMs as material rather than formal inference.

In addition to the fact that language model inference is material inference, it is also important that the formal logical relations that appear in LLMs are not directly coded into them. In the pre-LLM symbolist GOFAI, logical operators were embedded into the language processing system, making it difficult to represent everyday material relations. In contrast, LLMs acquire the material logical relations that exist between sentences from massive training data, and they mimic and output the logical relations that emerge therein. Logical inference in a language model is not encoded in the model, but is a byproduct of the weights of the neural network. It does not achieve classical logical inference perfectly. For example, researchers continue to face challenges in improving text entailment recognition tasks \autocite{PUTRA2024132,mccoy_right_2019,liu_inoculation_2019}, which ask whether there is a logical entailment relationship exists between sentences. In other words, the logic in LLMs is acquired bottom-up from massive training data, rather than top-down from formal logic implementation. In this sense, the subsymbolism \autocite{Chalmers2023-CHATCA-21} of LLMs is consistent with the logical expressivism of inferentialism.

Next, we will show that the quasi-compositional nature of LLMs is consistent with Brandom's quasi-compositionalism.

In the Transformer architecture, relationships with other tokens are realized through an attention mechanism, where vectors called \textit{attention} play a crucial role. Attention vectors take as arguments the matrix representing the input sequence and the vector of the target token itself, and form a probability simplex\footnote{A probability simplex is also called a probability vector. This refers to a vector whose components are non-negative and sum up to 1.} representing how strongly a given token depends on other tokens.

In light of the attention mechanism, which assigns weights to each token and forwards the attended representations to the next layer, it may seem that this supports the principle of compositionality. However, it remains unclear whether the principle of compositionality---namely, that ``the meaning of a complex expression is determined by its structure and the meanings of its constituents'' \autocite{sep-compositionality}---genuinely holds for such architectures. To investigate the problem of compositionality in LLMs, we must first examine the quasi-compositional character of inferentialism.

Inferential role semantics---including, but not limited to, Brandom's inferentialism---has long faced criticism for its apparent conflict with the principle of semantic compositionality. \textcite[hereafter F\&L]{Fodor1991-FODWMP} criticizes inferentialism for deviating from compositional principles of meaning. For example, inferential semantics would also admit the inference ``brown cows are dangerous'' as valid. But from a perspective that regards compositionality as essential to semantics, the meaning ``dangerous'' should be derived compositionally from the meanings of ``brown'' and ``cow.'' Since neither ``brown'' nor ``cow'' can yield the meaning ``dangerous,'' inferential semantics is considered inadequate as a semantics. As F\&L puts it, ```brown cows are dangerous' is contingently true, true in virtue of the facts about brown cows, not in virtue of the facts about meanings'' \autocite[24]{Fodor1991-FODWMP}.

In response, inferentialism, which understands meaning in terms of material inference, grounds meaning in the accumulation of normative material inferences within the discursive space. \textcite{d644c8a2-1203-361d-9d4f-40a6fcf6d285} notes that natural language is ``quasi-compositional,'' meaning that it is not fully compositional. The quasi-compositional stance of inferentialism maintains that ``[i]t is important not to treat languages as more compositional than they are'' \autocite[177]{brandom2007}. Czech inferentialist \textcite[chap.9]{InferentialismandtheCompositionalityofMeaning} claims that the principle of compositionality should not be understood as describing an actual act of composition, but merely as a methodological principle for decomposing the meaning of the whole into the meanings of its parts.

Now, let us turn to examining compositionality in LLMs. While there are various approaches to understanding compositionality in neural language models, we will focus specifically on how F\&L's criticism of inferential role semantics applies to LLMs. This focus reveals two crucial challenges to traditional compositional accounts when applied to LLMs.

The first challenge concerns how LLMs represent word meanings. The second, more fundamental challenge concerns how LLMs learn from text without distinguishing between different types of truth---specifically, between analytic truths that are supposedly true by virtue of meaning alone (like ``all bachelors are unmarried'') and synthetic truths that depend on empirical facts about the world (like ``brown cows are dangerous''). These two challenges work together to support a quasi-compositional view of LLM semantics.

The first challenge concerns word representation. Some might argue that LLMs exhibit compositionality based on famous experiments showing semantic relationships like $\vec{king} + \vec{woman} = \vec{queen} + \vec{man}$ \autocite{10.5555/2999792.2999959}. However, this argument fails to account for how modern LLMs actually work. These vector arithmetic relationships were discovered in static embedding systems like Word2Vec \autocite{mikolov2013efficientestimationwordrepresentations}, which assigned each word a single, fixed vector regardless of context. In contrast, modern Transformers create different representations for the same word in different contexts---what is called `dynamic embeddings.' The word `bank' receives different representations when discussing rivers versus finances. This context-sensitivity means that simple vector arithmetic cannot capture how LLMs compose meanings, making this compositional argument inapplicable to Transformer-based LLMs.

The second, more fundamental challenge emerges when we consider how F\&L's original criticism applies to LLMs. Recall that F\&L criticized inferential semantics for treating ``brown cows are dangerous'' as a valid semantic inference, arguing that this statement is only contingently true based on empirical facts about the world, not on the meanings of ``brown'' and ``cow'' themselves. F\&L's criticism assumes we can clearly distinguish between two types of statements: analytic statements like ``all bachelors are unmarried'' (true by meaning alone) and synthetic statements like ``brown cows are dangerous'' (true based on worldly facts).

This analytic/synthetic distinction, however, creates a fatal problem for applying F\&L's criticism to LLMs. Language models encounter both analytic and synthetic statements in their training data without any indication of which category they belong to. The model processes ``All bachelors are unmarried'' and ``Brown cows are dangerous'' through identical mechanisms---both appear as linguistic patterns to be learned from text. The model has no way to distinguish conceptual necessities from empirical contingencies, nor any means of accessing the ``worldly facts'' that supposedly ground synthetic truths.

This limitation reveals an important problem with F\&L's approach. F\&L's criticism relies heavily on maintaining the analytic/synthetic distinction---the idea that we can cleanly separate truths based on meaning from truths based on empirical facts. However, as \textcite{Quine1951-QUITDO-3} demonstrated in ``Two Dogmas of Empiricism,'' this distinction is far more problematic than it initially appears. Even seemingly clear cases like ``All bachelors are unmarried'' depend on linguistic conventions that could, in principle, be revised. What we take to be conceptual necessities are actually products of contingent linguistic practices within particular communities.

When we apply this Quinean insight to LLMs, an important consequence emerges. If the analytic/synthetic distinction cannot be cleanly maintained even in principle, then F\&L's criticism loses much of its force. LLMs treat all linguistic relationships as patterns derived from actual language use---exactly what we should expect if all semantic relationships are ultimately grounded in linguistic practice rather than a priori conceptual structures. In LLM training data, both ``All bachelors are unmarried'' and ``Brown cows are dangerous'' appear as instances of language use in context. Both derive their inferential power from the statistical patterns of how they have actually been used in discourse. From this perspective, there is no fundamental difference between these statements---both represent crystallized patterns of past linguistic behavior. The model learns that ``bachelor'' relates to ``unmarried'' through the same basic mechanism by which it learns that ``brown cows'' relates to ``dangerous'': through co-occurrence patterns and contextual associations in large-scale text data.

From this perspective, meaning determination in LLMs is essentially quasi-compositional. The meaning of individual lexical items is inductively constructed from countless examples of use in contexts in the training data, and that meaning is defined by inferential possibilities in new contexts. This is surprisingly similar to the structure of inferentialist semantics that Brandom advocates. In both cases, it is not analytic compositionality but the accumulation of past inferential patterns that produces meaning. The meaning processing of LLMs is therefore more compatible with the quasi-compositional character of inferentialist semantics than with the compositional principles assumed by traditional formal semantics.

\subsection{Substitution}\label{para:substition}

In inferentialism, singular terms and predicates are defined by means of inferential substitutions. However, LLMs do not introduce linguistic categories such as singular terms or predicates by design. These categories are acquired automatically and verified post hoc (In contrast, in formal semantics such as Montague grammar, singular terms and predicates are given a priori as entirely different linguistic categories). LLMs do not extract singular terms by means such as ``if $Qa \implies Qb$ and $Qb \implies Qa$, then $a$ and $b$ have the same meaning as singular terms.'' The same applies to predicates. Thus, there is a mismatch between the inferentialist approach of extracting singular terms and predicates via substitutional inference and the way singular terms and predicates are handled within LLMs.

However, there might be a connection between the role of substitutability in generating meaning or content \autocite{sep-information} and the process by which LLMs produce tokens according to induction heads \autocite{olsson2022context} and suppression heads \autocite{mcdougall-etal-2024-copy}. Induction heads and suppression heads refer back to previous tokens and replicate or suppress them. On the other hand, inferentialism focuses on normative linguistic practices based on non-commutativity, such as the `material incompatibility' that one cannot simultaneously commit oneself to both ``this apple is green'' and ``this apple is red'' \autocite{ae1d0a0a-8dea-323c-89d0-72c600931595,Brandom1994-BRAMIE}. The role of heads that deal with substitutability, such as induction or suppression heads, may be key to explaining normativity in LLMs in a way that does not fall into the dispositionalism that inferentialism criticizes. Their properties might be crucial to a non-dispositional account of normativity that inferentialism seeks.

\subsection{Anaphora}\label{para:anaphora}

The inferentialist way of resolving concepts like deictics and reference through anaphora treats the meaning of demonstratives not in relation to the world, but rather internally within the language. This approach is consistent with the anti-representational nature of LLMs. In inferentialism, the relation to the world is ultimately given, and demonstratives are given by substitutional inference involving anaphora of terms within sentences. Non-multimodal LLMs, which take no input other than linguistic data, cannot connect to the world through non-linguistic media, and so concepts such as deixis and reference remain trapped in language.

\begin{figure}[htbp]
    \centering
    \includegraphics[width=0.8\textwidth]{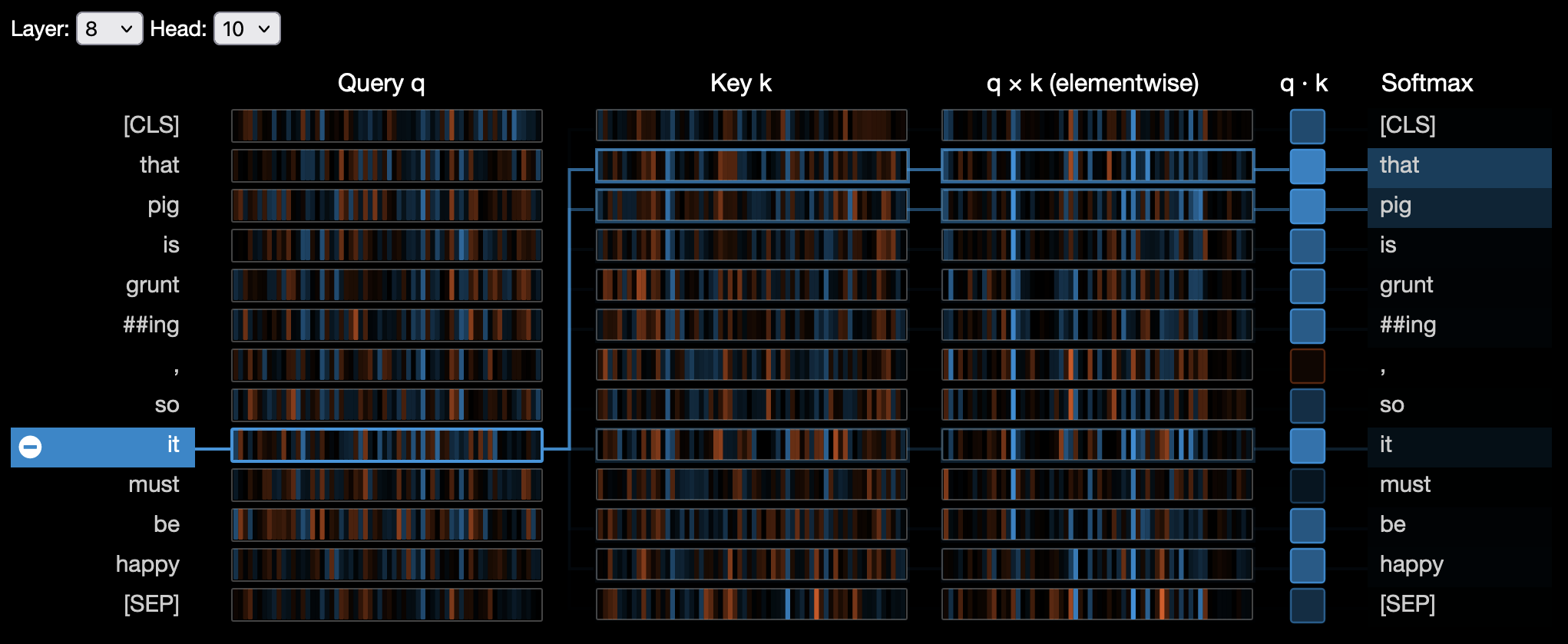}
    \caption{Neuron View by BertViz. The inner product between ``that'' and ``pig'' is large respectively, showing anaphora resolution of a demonstrative. The color and brightness of each neuron represent the sign (blue for positive, red for negative) and magnitude of the inner product value of each neuron, respectively.}
    \label{fig:bertviz}
\end{figure}

In Transformers, anaphora can be considered to be realized by the attention mechanism or by special attention heads known as `induction heads.' \footnote{
Masked attention is used to prevent backward references in Transformers. The encoder model BERT \autocite{devlin-etal-2019-bert} utilizes self-attention, permitting backward references, while the decoder model GPT \autocite{yenduri2023generativepretrainedtransformercomprehensive} employs masked attention that prohibiting backward references.
} Attention represents a probability simplex, denoting the extent to which one token is dependent on other tokens. \cref{fig:bertviz} visualizes the attention allocated by the demonstrative ``it'' in the sentence ``That pig is grunting, so it must be happy'' within BERT, one of the Transformer encoder models\footnote{
This figure was created using BertViz \autocite{vig-2019-multiscale}, displaying Neuron View on the \texttt{bert-base-uncased} pretrained model. The figure shows the attention in layer 8, head 10.
}. As seen in the figure, the demonstrative ``it'' exhibits a substantial inner product (similarity) with both ``that'' and ``pig,'' indicating that BERT is engaged in anaphora resolution of the demonstrative.

Induction heads are circuits that, given a sequence ``[A][B]...[A],'' select ``[B]'' as the subsequent token to complete the pattern by replicating previously occurring sequences \autocite{elhage2021mathematical,olsson2022context}. To illustrate, when presented with a token sequence ``That pig is grunting ... it'', and with attention linking the demonstrative and its referent, the induction heads would have a heightened probability of selecting the word ``grunts'' next.

By means of attention and induction heads, demonstratives are associated with their referents, and the surrounding context is also linked. Since the LLM can only imagine the world through the linguistic information given to it, the meaning of demonstratives such as ``this pig'' or ``it'' is not obtained by pointing to the world, but through forward anaphora resolved by attention.

\section{Anti-Representationalist Nature of Large Language Models}\label{sec:llm_anti_representationalism}

In this section, we point out the anti\hyphen{}representationalist and linguistically idealist characteristics of LLMs and, on that basis, discuss their semantic properties and affinity with semantic internalism.

\subsection{The Anti-Representationalism of Large Language Models}\label{subsec:llm_anti_representationalism}

As seen in the previous chapter, LLMs exhibit strongly anti\hyphen{}representationalist characteristics in their learning processes and functionalities. The only points at which one might consider LLMs to be \emph{connected} to the world are:

\begin{enumerate}
    \item Weight adjustments through training data,
    \item RLHF (DPO, etc.), and
    \item In-context learning that occurs during model-human interaction.
\end{enumerate}

Weight adjustments are made based on large-scale corpus data, RLHF is a reinforcement learning-based method that uses human evaluations of the model's responses to adjust internal parameters \autocite{ouyang2022traininglanguagemodelsfollow,kaufmann2024surveyreinforcementlearninghuman}, and in-context learning refers to parameter shifts during ongoing interaction with the model \autocite{dong-etal-2024-survey}. The corpus data and human evaluations used for RLHF, as well as the interactive inputs for in-context learning, can all be freely altered by humans. Thus, there is no principled guarantee that the system maintains a representational relation with the world. For example, consider replacing every instance of the sentence

\begin{quote}
Apples are red.
\end{quote}

in the training data with

\begin{quote}
Apples are rainbow-colored.
\end{quote}

In this scenario, the LLM would likely come to \emph{believe} that ``apples are rainbow-colored.'' Likewise, if the model's ethical responses are consistently rated poorly during RLHF, the model is likely to begin producing unethical responses. In this way, LLMs are \emph{only} open to the truths of the world via the humans involved in providing training data and RLHF, and these truths can be easily distorted by humans. Indeed, one might say that distributional semantics itself is fundamentally anti-representationalist. \textcite{10.1007/978-3-030-34974-5_14} states:

\begin{displayquote}[{\cite[14]{10.1007/978-3-030-34974-5_14}}]
Compositional symbolic semantics is usually referential, and presupposes reality outside language, [...] and it may be seen as realism. Contextual distributional semantics does not presuppose anything outside language, deriving meaning vectors just from contexts within language. It may thus be seen as antirealism.
\end{displayquote}

These characteristics of LLMs allow us to say that they are anti\hyphen{}representationalist. Furthermore, non-multimodal, purely linguistic LLMs examined in this study can be regarded as linguistically idealist in the sense that they are devoid of any perceptual experience. The phenomenon known as \textit{factuality hallucination}---where LLMs generate entities or relations that diverge from actual world facts---is also consistent with the anti-representationalist nature of LLMs discussed here \autocite{Huang_2025}. \textcite{Rorty1979-RORPAT-3} contended that the anti-representationalism espoused by Sellars, Quine, and Davidson, respectively, stemmed from their critiques of the given myth, the distinction between analysis and synthesis, and the conceptual frame. Given that the non-multimodal LLMs are devoid of sensory sensors, the issue of grounding in sense data is not a concern, nor is the argument of a given myth. The anti-representational nature of LLMs is rooted in their linguistic idealism, irrespective of the approach adopted by Rorty. That is, even irrespective of Rorty's argument, the linguistic idealism intrinsic to non-multimodal LLMs---models devoid of perceptual grounding---suffices to substantiate their anti-representationalist characterization.

In LLMs, notions such as truth and reference are not treated as ontological entities existing outside language, but rather as internal components of discursive practice. This deflationary view aligns with the inferentialist position known as truth expressivism, in which a statement like ``P is true'' is not a metaphysical claim about correspondence with the world but simply expresses the speaker's commitment to the proposition P \autocite[chap.5]{Brandom1994-BRAMIE}. When an LLM outputs the statement ``P is true,'' it should not be interpreted as asserting a transcendent or correspondence-based truth. Instead, it merely indicates the LLM's internal state of commitment to P within the ongoing discourse, such as a conversation with a human. Truth predicates are nothing other than expressions of the LLM's beliefs. Thus, the internalist and deflationary treatment of truth in inferentialism can be seen as fundamentally consistent with the anti-representationalist nature of LLMs.

\section{Truth and Normativity}\label{sec:llm_truth}


In this section, I examine theories of truth and normativity in LLMs. In Brandom's inferentialist framework, whether a proposition is true is determined not by ultimate, transcendent, or fixed factors, but rather by the normative attitudes and normative statuses among language users, with its concrete implementation theorized as scorekeeping (the practice whereby participants in a language game mutually record and update each other's entitlements and commitments to speech acts). This theory of truth that determines truth within a community of language users is generally called the \textit{consensus theory of truth}, including, besides inferentialism, such theories as \textcite{Habermas1991-HABCAT-2}'s universal pragmatics. The distinctive feature of this consensus theory is that, in contrast to the realist correspondence theory of truth, it guarantees the inevitable epistemological consequence of Kantian idealism regarding the knowing subject while also accounting for the public nature of truth.

As discussed in the previous section, due to the arbitrariness of training data in LLMs' architecture, their internal models (representations) do not necessarily exist solely as representations of a realist world. Moreover, lacking sensory organs, they cannot adopt a foundationalist theory of truth. LLMs are effectively disconnected from the truth or falsity of facts in the world and, without sensory organs, cannot adopt conceptualism. LLMs are precisely in what \textcite[11]{McDowell1996-MCDMAW-7} calls a state of `frictionless spinning in a void.' Therefore, a theory of truth for LLMs must be constructed with consideration of these characteristics.

A key factor in considering a theory of truth for LLMs is the \textit{public} nature of conversational LLMs such as ChatGPT and Claude. \textcite{Misak2013-MISTAP}, who belongs to the same pragmatist tradition as Brandom, opposes the linguistic picture of correspondence between the world and sentences (propositions), stating:

\begin{displayquote}[{chap.~3.5}]
For the very idea of the believer-independent world, and the items within it to which beliefs or sentences might correspond, seems graspable only if we could somehow step outside our corpus of belief, our practices, or that with which we have dealings. [...] The correspondence theory makes truth ``a useless word'' and ``having no use for this meaning of the word `truth', we had better use the word in another sense'' (\cite[\nopp 5.553]{Peirce1931-PEICP-2}, as cited in Misak)
\end{displayquote}

To prevent the concept of \textit{truth in LLMs} from becoming meaningless, we need to adopt not a correspondence theory of truth or an idealist/solipsistic theory of truth (e.g., \textcite[chap.~10]{Jackendoff2002-JACFOL-2}), but rather a theory that emphasizes the public aspect of dialogue between LLMs and humans. The neo-pragmatist lineage descended from Rorty emphasizes this aspect of publicity among linguistic subjects over metaphysical conceptions of truth. Brandom's inferentialism, in particular, is a theory that grounds semantics in pragmatics among linguistic subjects. As we saw in the previous chapter, mechanisms such as RLHF, DPO, and prompting realize interaction between humans and LLMs, and this public aspect of conversational LLMs aligns with the inferentialist conception of truth.

Truth in inferentialism is based on the notion of normativity among linguistic subjects, and what counts as correct ultimately follows the dynamics of norms in the linguistic practices to which those subjects belong. However, Brandom himself takes a cautious stance toward naturalistically reducing normativity \autocite[pp.~42--6]{Brandom1994-BRAMIE}. In response, there are approaches that attempt a naturalistic explanation of normativity while preserving the basic insights of inferentialism, such as \textcite{Sapalski2025-SAPAQA}'s approach that draws on the conceptual distinction between internal and external points of view found in \textcite{Hart2012-HARTCO-112}. According to her research, the perspective from within a community on the same norm differs from the external perspective. That is, while subjects belonging within a language game follow the norms of the language game itself, from outside the language game, this can be understood as dispositions of the language game itself. Brandom states that dispositionalism (regularism) is inadequate as an explanatory item for norms because ``they may fail to make room for the crucial distinction between [...] what is done and what is correct or ought to be done'' \autocite[41]{Brandom1994-BRAMIE}. However, if we introduce the internal/external perspective distinction into inferentialism, it seems possible to explain normativity in a naturalistic manner\footnote{An attempt to naturalize inferentialism from an evolutionary and biological standpoint can be found in \textcite{Peregrin2022-PERIN}. However, given that LLMs lack an evolutionary history, the application of this approach to them poses significant challenges.}.

Conversational LLMs can receive human evaluations of their output sentences---for example, ratings as \textit{good} responses or \textit{bad} responses---through mechanisms like RLHF. More precisely, humans provide preference feedback by comparing multiple outputs to determine which is more appropriate, and a reward model is learned from this preference data. The language model then continues updating its internal parameters to maximize the continuous reward signal output by this reward model. Let us observe this process through the lens of normativity. For a language model, whether its output sentences are correct can be interpreted as determined in light of feedback provided by humans. Truth is established in this interaction between the language model and humans. The problem of normativity concerns whether a claim or sentence is correct or incorrect; correct claims are claims that ought to be made, while incorrect claims are claims that should not be made. Here, the correctness of sentences that conversational LLMs \textit{ought} to output is grounded in the preference feedback provided by humans and the reward model based on it. Negative feedback acts as a form of normative `sanctioning' \autocite[chap.~1]{Brandom1994-BRAMIE}. That is, correctness is established within the discursive practices shared between conversational LLMs and humans. Furthermore, conversational LLMs today are not limited to interacting with humans. In an architecture known as Reinforcement Learning from AI Feedback (RLAIF), LLMs can evaluate and reinforce each other's outputs through interaction \autocite{10.5555/3692070.3693141}. In this discursive space, no humans are present. Needless to say, this normative system is ontologically naturalistic (the internal states of LLMs are describable as physical computational processes; \textit{residual stream} is natural kind).

In this way, the compatibility of normativity grounding inferentialism with naturalism is demonstrated through the concrete case of LLMs. Normative practices in human-language model interaction, while implemented as physical and computational processes, enable meaningful engagement with normative concepts such as truth and correctness. Therefore, it can be concluded that a consensus theory of truth that emphasizes this public and dialogical character is most suitable as a theory of truth for conversational LLMs.

\section{Limitations in the Inferentialist Interpretation of LLMs}\label{sec:tensions}

While the inferentialist framework offers valuable insights for understanding LLMs, several fundamental tensions emerge between the philosophical commitments of inferentialism and the nature of large language models. These tensions do not necessarily invalidate the inferentialist approach, but they highlight areas requiring further theoretical development.

\paragraph{Rationalism and the Sapient-Sentient Distinction.} Brandom inherits the rationalist spirit from Kant, particularly in his distinction between sapient and sentient beings. According to Brandom, sapient creatures differ from merely sentient ones by employing concepts within inferential reasoning, or within the space of reasons \autocite[chap.~1]{Brandom1994-BRAMIE}. This raises a fundamental question: on what grounds can LLMs be considered rational agents capable of sapient behavior? The question of whether LLMs genuine conceptual employment in inferential contexts requires systematic investigation beyond the scope of this paper.

\paragraph{Propositionalism and Sub-symbolic Processing.} Inferentialism is fundamentally committed to propositionalism \autocite{blunden_robert_2012}. Brandom explicitly advocates for the ``pragmatic priority of the propositional'' \autocite[79]{Brandom1994-BRAMIE}, and logical inferentialism is a framework which develops material inference within discrete propositional structures. The substitutional inference framework introduced in this paper similarly relies on the discreteness of symbolic substitution. However, LLMs are what operate as sub-symbolic systems, processing continuous vector representations rather than discrete propositional units. This creates a fundamental mismatch between inferentialism's commitment to propositional discreteness and LLMs' continuous processing architecture.

There are two opposing approaches concerning this tension. Eliminativists such as \textcite{Churchland1981-CHUEMA-2} argue that if natural kinds corresponding to propositions cannot be identified within neural networks, the concept of proposition should be abandoned in favor of purely connectionist explanations. Conversely, recent work on \textit{propositional interpretability} \autocite{ChalmersManuscript-CHAPII-5} attempts to interpret connectionist systems with propositional concept. Whether this tension can be resolved without abandoning either propositionalism or the sub-symbolic nature of LLMs remains an open theoretical challenge.

\paragraph{The Privileging of Assertion.} \textcite{Kukla2009-KUKYAL,Rouse2015-ROUATW-2} criticizes inferentialism for privileging assertion as the fundamental speech act type, arguing that this privileging reflects unexamined assumptions about the nature of linguistic practice (\textcite[chap.~1]{Kukla2009-KUKYAL} characterizes the privileging of declarative sentences in Brandom's inferentialism as a ``serious error,'' referring to it as the ``declarative fallacy.''). In the context of LLMs, it is unclear whether their architectural and functional organization exhibits the kind of systematic differentiation between assertion and other speech act types that inferentialism presupposes (LLMs would likely have no trouble handling vocatives such as `Yo!' or `Lo!').

\paragraph{Conceptual Realism.} Brandom's inferentialism adopts a form of conceptual realism, holding that ``the way the world objectively is, in itself, is conceptually articulated'' \autocite[54]{Brandom2019-BRAASO-13}. This position has been subjected to penetrating criticism by \textcite{Habermas2003-HABTAJ-3}, who argues that insofar as inferentialism adopts conceptual realism, ``by making claims about the structure of the world `in itself,' undermines his discourse-theoretic analysis of reality as `appearing' in language'' \autocite[152]{Habermas2003-HABTAJ-3}.

This criticism aligns with a central thesis of this paper: that LLMs are best understood as systems that say nothing about the structure of the world, and that concepts of truth and meaning emerge productively only within the interactive space between language models and their interlocutors--—whether human or artificial. The anti-representationalist interpretation of LLMs developed here suggests that their linguistic behavior should be understood without reference to world-directed representational content. This position is consistent with Habermas's critique of conceptual realism, as both reject the notion that successful linguistic practice requires an independently conceptually-structured reality. Within inferentialism, the combination of the aforementioned propositionalism and the consensus theory of truth leads to a worldview in which the world is conceived as the totality of propositional facts. It remains to be examined whether this worldview has the potential to be reconciled with the anti-representationalist interpretation of LLMs proposed in this paper.

\section{Conclusion}

In this paper, we have attempted to interpret and describe the behavior of current Transformer-based LLMs through Robert Brandom's inferential semantics, which is representative of the analytic-pragmatist tradition.

The subjective nature of current deep learning models and LLMs can be seen as reinforcing the position of `subject naturalism' as proposed by \textcite{Price2011-PRINWM}. His subjective naturalism, influenced by Brandom, does not prioritize the explanation of the world (the object) through the lens of natural science---a concept referred to as `object naturalism.' Instead, it places emphasis on the explanation of human linguistic and cognitive practices (the subject) from a naturalistic standpoint. This position is analogous to the inferentialist approach adopted in this paper, as it is characterized by an attempt to elucidate the function of language not as a representation of the world, but from within the discursive practices themselves. As ontologically naturalistic language models begin to exhibit subjectivity, traditional physics-centered naturalism may be subject to further development.

This paper represents one philosophical attempt to interpret rapidly evolving AI technologies, especially LLMs. Future challenges include empirically testing the theoretical framework presented here and exploring new philosophical problems that may arise as LLMs continue to evolve. We also anticipate that the findings of this study will provide new perspectives on AI ethics and the coordination between humans and AI.

\printbibliography[
  title=References,
]

\end{document}